\documentclass[sigconf,nonacm]{acmart}

\usepackage{makecell}

\AtBeginDocument{%
  \providecommand\BibTeX{{%
    \normalfont B\kern-0.5em{\scshape i\kern-0.25em b}\kern-0.8em\TeX}}}

\acmDOI{10.1145/nnnnnnn.nnnnnnn} 
\acmISBN{978-x-xxxx-xxxx-x/YY/MM} 
\acmConference[GECCO '21]{The Genetic and Evolutionary Computation Conference 2021}{July 10--14, 2021}{Lille, France}
\acmYear{2021}
\copyrightyear{2021}

\begin{document}

\title{GEVO-ML: Optimizing Machine Learning Code with Evolutionary Computation}

\author{Jhe-Yu Liou}
\affiliation{%
   \institution{Arizona State University}
   \city{Tempe}
   \state{AZ}
   \country{USA}}
\email{jhe-yu.liou@asu.edu}
\author{Stephanie Forrest}
\affiliation{%
   \institution{Arizona State University}
   \city{Tempe}
   \state{AZ}
   \country{USA}}
\affiliation{%
   \institution{Santa Fe Institute}
   \city{Santa Fe}
   \state{NM}
   \country{USA}}
\email{stephanie.forrest@asu.edu}
\author{Carole-Jean Wu}
\affiliation{%
   \institution{Arizona State University}
   \city{Tempe}
   \state{AZ}
   \country{USA}}
\email{carole-jean.wu@asu.edu}

\begin{abstract}

Parallel accelerators, such as GPUs, are key enablers for large-scale Machine Learning (ML) applications. However, ML model developers often lack detailed knowledge of the underlying system architectures, while system programmers usually do not have a high-level understanding of the ML model that runs on the specific system. To mitigate this gap between two relevant aspects of domain knowledge, this paper proposes GEVO-ML, a tool for automatically discovering optimization opportunities and tuning the performance of ML kernels, where the model and training/prediction processes are uniformly represented in 
a single intermediate language, the Multiple-Layer Intermediate Representation (MLIR).
GEVO-ML uses multi-objective evolutionary search to find edits (mutations) to MLIR code that ultimately runs on GPUs, improving performance on desired criteria while retaining required functionality.  

We demonstrate GEVO-ML on two different ML workloads for both model training and prediction. GEVO-ML finds significant Pareto improvements for these models,  
achieving 90.43\% performance improvement when 
model accuracy is relaxed by 2\%, from 91.2\% to 89.3\%. For the training workloads, GEVO-ML finds a 4.88\% improvement in model accuracy, from 91\% to 96\%, without sacrificing training or testing speed. Our analysis of key GEVO-ML mutations reveals diverse code modifications, while might be foreign to human developers, achieving similar effects with how human developers improve model design, for example, by changing learning rates or pruning non-essential layer parameters.

\end{abstract}


\maketitle

\section{Introduction}
\label{sec:introduction}

Machine learning (ML) applications are being deployed at unprecedented scales across a wide variety of domains.  
This is enabled by advances in many research domains cutting across the entire system stack---from domain-specific programming frameworks, such as PyTorch~\cite{paszke2017automatic}, TensorFlow~\cite{tensorflow} to domain-specific architecture support and hardware accelerators, such as Nvidia Graphic Processing Units (GPU) with Tensor Core~\cite{nvvolta}, Intel CPUs with AVX-Vector Neural Network Instructions~\cite{intelvnni}, Tensor Processing Units (TPU)~\cite{tpu}. 
To achieve faster model training and prediction execution time while improving model accuracy, application model developers fine-tune many of design components. 
For example, in the network architecture layer, developers must decide how many layers will be in the network, how many neurons will be in each layer~\cite{stathakis2009many}. Similarly, in Support Vector Machine (SVM), designers must select a cost value~\cite{hsu2003practical}. And, at the framework level, developers select the operators to be used in the model and allocate available system resources.

As the scope of artificial neural networks has grown, the raw number of tunable knobs has exploded. Most designers use empirical methods to identify a set of parameters that works well for the respective machine learning tasks. Prior research has examined automated parameter tuning methods, or hyperparameter search, to find  model parameters~\cite{bergstra2012random, Thornton:2013:ACS:2487575.2487629}. These include simple grid search~\cite{Larochelle:2007:EED:1273496.1273556}, random search~\cite{bergstra2012random}, reinforcement learning~\cite{zoph2016neural, bello2016neural}, evolutionary computation (EC)~\cite{neat}, and gradient descent~\cite{liu2018darts}.

At the framework level, developers have introduced ML compilers, which 
find optimizations such as operator fusion~\cite{operator_fusion}, or at code generation time can determine the degree of loop unrolling or loop tiling based on the hardware platform characteristics~\cite{chen2018tvm,ragan2013halide}. These searches and optimizations occur at different levels and are usually performed separately.
There are a number of other framework-level features, such as threading libraries and scheduling policies that can be used to further optimize ML training and inference execution time~\cite{Wang:arxiv19,inteltips,tensorflowperformance}.

Typically, developers leave low-level CPU code optimization to compilers, which for most applications is hard to beat. However, GPU codes are often much harder to optimize, thus, leaving ample rooms for performance improvement. Unless the developer has detailed knowledge of the underlying system architecture(s), it is challenging to uncover these additional optimization opportunities. 
Earlier work on GEVO showed that EC can find many interesting GPU optimization opportunities, reporting an average 49\% speedup on common general-purpose parallel benchmarks~\cite{GEVO.TACO}.  These speedups are achieved by relaxing the usual compiler requirement to preserve exact program semantics. This approach is thus well-suited for approximate computing applications such as ML.
GEVO can be thought of as a compiler post-pass performance tuning framework, which encodes optimization objectives, such as execution time, energy use or accuracy, in its fitness function and implements a set of mutation and recombination operators for GPU kernel transformations in the LLVM intermediate representation (LLVM-IR).

Although the approach described in GEVO is general in the sense that it can be applied to any GPU programs represented in LLVM-IR, it is well-suited to ML workloads that are computationally-intensive and error-tolerant. Indeed, one neural network example is included in the earlier work. 
However, GEVO has limitations in this domain.  For instance, it can only optimize certain model layers or operations---those that can be compiled into LLVM-IR. In many ML frameworks most operations are implemented to invoke the device vendor library where the source code and LLVM-IR are not available. This restriction drastically limits which part of a neural network model can be tuned by GEVO. Further, without access to the high-level neural network model architecture, GEVO could discover only local optimizations within a single neural network operation. 

This paper presents GEVO-ML, an EC approach for optimizing ML workloads, which addresses the aforementioned issues. 
First proposed in~\cite{gevoml_proposal},
GEVO-ML optimizes ML models expressed in the Multi-Level Intermediate Representation (MLIR), which represents the entire model in a single representation. Moving to the MLIR representation required designing new mutation operators and overcoming significant engineering challenges. Our evaluation shows that GEVO-ML finds optimizations both to the ML model 
and to the low-level code implementation.

To summarize, the key contributions of this paper are:
\begin{itemize}

    \item We present GEVO-ML---an EC-based tool for finding cross-layer optimizations for machine learning workloads that are expressed in the MLIR format.  GEVO-ML generalizes across ML models, MLIR dialects, and underlying system architectures. It finds Pareto solutions for different tradeoffs of model accuracy and runtime and identifies optimizations that are tailored to the particular workloads (Section~\ref{sec:Design}).
    
    \item With a novel mutation operator that resizes tensor variables, GEVO-ML uncovers optimization opportunities, ranging from model architectures (e.g., by removing unneeded network layers) to low-level implementation inefficiencies, by leveraging nuanced interactions across abstraction layers. We follow up the evaluation for GEVO-ML with in-depth code analysis (Sections~\ref{sec:result-mobilenet} and ~\ref{sec:result-2fc}). 
    \item We perform detailed experimental evaluation to assess the performance of GEVO-ML, using two different ML models: MobileNet on the CIFAR10 dataset~\cite{cifar10} and a two-layer fully-connected neural network on the MNIST~\cite{mnist} dataset. For MobileNet as the model prediction task, GEVO-ML finds optimizations that achieve 90.43\% performance improvement at a cost of 2\% model accuracy.  For the model training task on the two-layer fully-connected neural network, GEVO-ML finds optimizations that improve model accuracy by 5\% without changing the runtime performance (Section~\ref{sec:result}).
    
\end{itemize}

We plan to open source GEVO-ML~\emph{URL-elided-for-blind-review} upon paper acceptance to advance the field with open and reproducible science.

\section{Related Work}

\label{sec:related_work}

Automating the construction and optimization of machine learning models, known as AutoML, is a growing body of research. An array of search and optimization methods in this domain includes evolutionary computation, reinforcement learning, and superoptimization. 
Using evolutionary computation (EC) to improve ML workloads dates back to 1989, where Montana and Davis proposed using EC to train a neural network~\cite{montana1989training}.
The most established and commonly-used approach in this domain is NEAT, first proposed by Stanley et al. in 2002~\cite{neat}, which simultaneously learns the connection topology and weights for each neuron.  Since then, many papers have expanded the NEAT approach to operate on larger networks and more complex tasks~\cite{stanley2009hypercube, verbancsics2011constraining, real2017large, CoDeepNEAT}. 

More recently, convolution neural networks (CNNs) have achieved extraordinary performance in image classification tasks by providing additional convolution layers as filters. 
These layers are used to identify relevant spatial patterns in images so the number of features can be reduced before being fed into a traditional neural network. Many approaches, including applying reinforcement learning, for identifying performant CNN architectures (topologies) have been proposed~\cite{pham2018efficient, bender2018understanding, liu2018progressive, kandasamy2018neural, liu2017learning, liu2017hierarchical, xie2017genetic, zoph2018learning}, outperforming manually designed architectures on several tasks. Similar to NEAT, Real et al. proposed using EC to design CNNs in a limited search space of convolution layers composed by common arithmetic operations~\cite{real2019regularized}. This work achieves state-of-the-art performance classifying the ImageNet dataset compared to other network architecture searches, which use random search and reinforcement learning. NSGA-Net~\cite{nsganet} compliments the work of Real et al. by seeking to improve model accuracy as well as decreasing the model complexity, using a well-known, multi-objective selection method in EC---NSGA-II~\cite{nsga-ii}. 

The aforementioned prior works with EC as the search method have one thing in common---they rely on customized binary encoding, usually corresponding to the connection of selected, common operations used in CNN architecture. The encoding serves as the individual representation for EC to search. 
In a lower level of the system stack, Liou et al. proposed GEVO~\cite{GEVO.TACO}, that searches the implementation of common neural network operations, in particular, the stochastic gradient decent operation, in the form of LLVM intermediate representation. While GEVO targets low-level code, the result implies that the discovered optimization has high-level intention similar to weight pruning techniques.
Our work, GEVO-ML, based on GEVO, extends the individual representation to the entire neural network model ,through multi-level intermediate representation.

The most comprehensive work that is comparable to this paper is AutoML-Zero~\cite{automl-zero}.
AutoML-Zero engages the fundamental mathematical operations as the building blocks to compose basic ML tasks: training and prediction. Starting from almost scratch, AutoML-Zero utilizes EC search and eventually rediscovers the algorithm similar to stochastic gradient decent for the purpose of training. The results showcase that EC search within a generic framework can discover human knowledge with minimal human intervention and restriction. While the result is insightful and impressive, searching from scratch is extremely resource hungry. Around 50,000 CPU days (10,000 processors for 5 days) are required. GEVO-ML intends to search based on an existed and established algorithm or a neural network model and looks for opportunities for performance improvement. While fundamental designs are alike between AutoML-Zero and GEVO-ML, GEVO-ML leverages a more standard, cross-board representation from compiler experts instead of customized and selected operations. As a result, GEVO-ML is easier to deploy in a production-ready environment and can perform code optimization search for machine learning models described in any machine learning framework, running on a wide array of hardware back-ends, including GPUs. 

As introduced above, with search on top of code representation, 
our work and relevant prior works also relate to program synthesis, where superoptimization provides as an alternative option as the search process. Unlike EC which relies on pre-defined test cases for verification, superoptimization transforms a given program into a boolean equation and search for the program rewrite where semantic equivalence is guaranteed. Thus, test cases are often not required for validation. Jia et al. recently proposed TASO~\cite{taso}, which uses superoptimization methods to optimize a computational graph of a deep neural network. TASO enumerates all possible combinations of operator implementations and selects the graph implementation that minimizes runtime.  A SAT solver is used to ensure that the original graph's functionality is preserved.  Although promising, this approach currently does not 
scale beyond small graphs comprised of more than four operators.

\section{Background}

GEVO-ML's design targets the Multi-level Intermediate Representation (MLIR) representation for Deep Neural Networks (DNNs). This section discusses relevant representations of MLIR and reviews how MLIR fits into the TensorFlow deep learning framework.

\subsection{LLVM Intermediate Representation}
The core of the LLVM compiler is its intermediate representation, LLVM-IR~\cite{LLVM:CGO04}. LLVM-IR is a strongly-typed, abstract assembly language that is target-independent. The LLVM compiler front-end first compiles high-level languages like C/C++ into this IR and applies code optimization without considering target device dependence. 

LLVM-IR uses Single Static Assignment (SSA) with infinite register allocation to enable many modern compiler optimizations, including data flow and variable reachability analysis, dead code elimination, and other optimizations we expect from modern compilers. Many projects built on top of LLVM use both the flexible extensibility of SSA and the surrounding LLVM compiler infrastructure. Relevant examples include: Nvidia NVVM~\cite{nvvm} which is an extension to LLVM-IR for their GPU code (CUDA) compilation; Glow~\cite{rotem2018glow} which is a compilation framework for PyTorch DNN models; and, GEVO which is is built on top of LLVM-IR and Nvidia NVVM. 
As the number of LLVM-IR extensions for DNNs grow, LLVM developers realized that 
similar functionality was being developed repeatedly for different extensions with similar domains.  MLIR attempts to unify these efforts. 
\begin{figure}
    \includegraphics[width=1\linewidth]{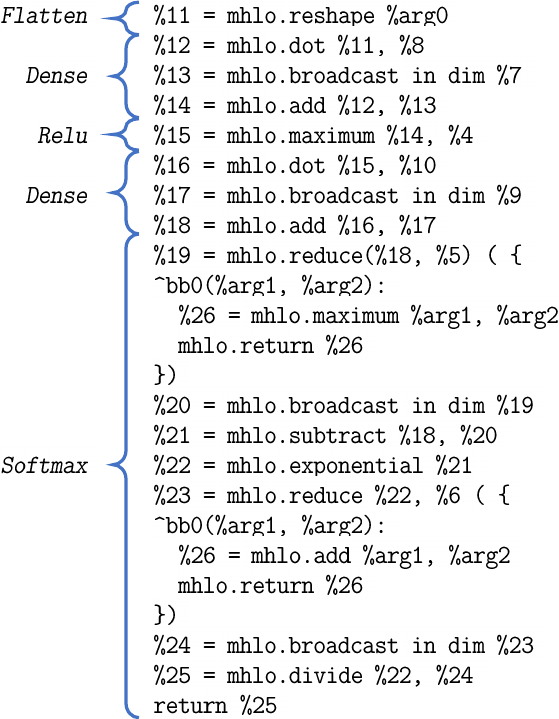}
	\caption{A MLIR code snippet from a neural network with two fully-connected layers, written in TensorFlow. The high level code is presented on the left while the translated MLIR code is shown in the HLO dialect on the right.
	The operand type and the attribute in each MLIR operation are omitted for clarity.}
	\label{fig:2fc_mhlo} 
\end{figure}

\subsection{Multi-Level Intermediate Representation (MLIR)}
Mature deep learning frameworks, such as TensorFlow, allow programmers to conveniently construct a deep neural network as a sequence of high-level operations using convolution, fully connected layers, and such. The model is then translated into the the intermediate representation 
and further optimized by the framework compiler.

\begin{figure*}[t]
	\includegraphics[width=1\textwidth]{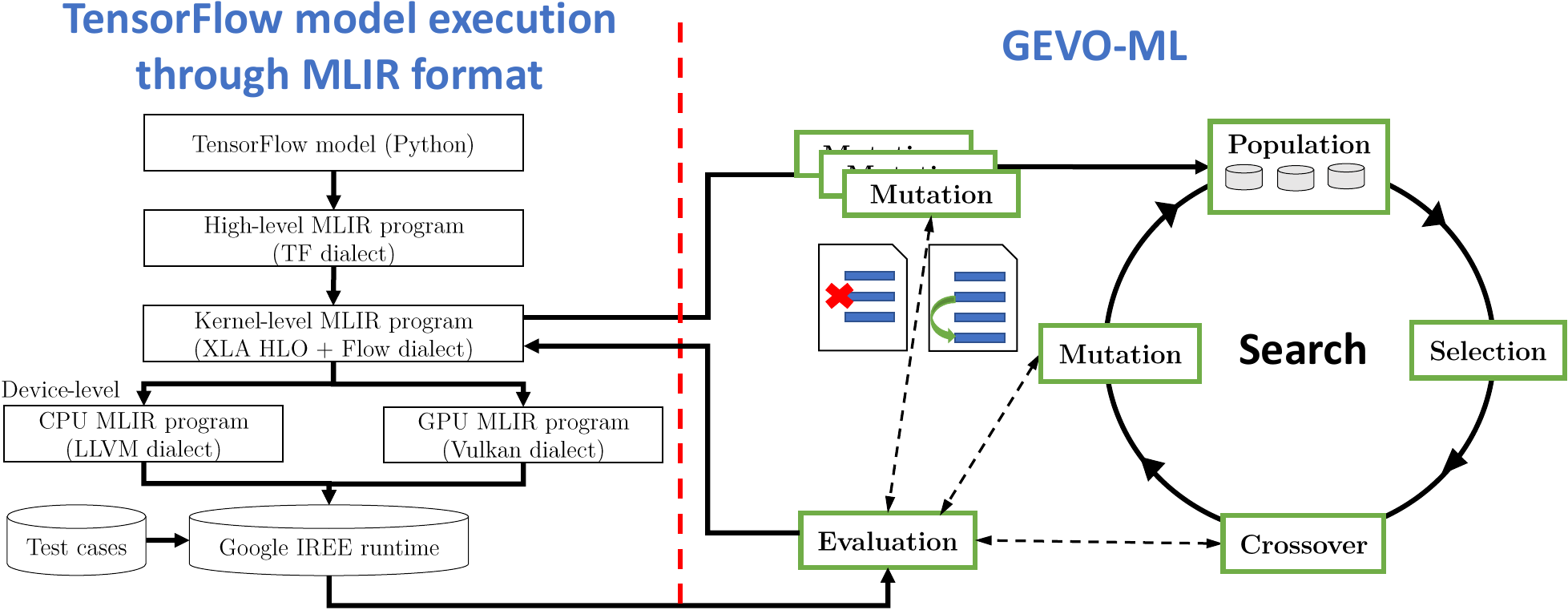}
	\caption{GEVO-ML execution flow in the context of a TensorFlow model running under Google's IREE environment.}
	\label{fig:gevoml} 
\end{figure*}

Eventually, the instructions are mapped to a device, like a GPU, which executes the model.  Many compiler optimizations can be performed at each step of this multi-level compilation process. These optimizations are applied to abstraction layers across the different frameworks, compilers, and hardware systems. There are many redundancies in this process. For example, within one framework, same optimization, such as dead code elimination, can be applied repeatedly across different abstract layers. Another example is that, when spanning across different frameworks, DNN optimizations are essentially linear algebra domain optimizations, and they appear in many established frameworks, including 
Intel's nGraph~\cite{cyphers2018intel} and image processing frameworks such as Halide~\cite{ragan2013halide}, as being reinvented. To address these redundancies, 
Lattner et al. proposed MLIR~\cite{mlir} 
to enable developers representing different abstraction levels in a customized operation or instruction set under a unified compiler infrastructure.

The centerpiece of MLIR is dialects. A MLIR dialect is a developer-defined and customized operation set (operations in MLIR are similar to instructions in low-level representation). Despite customization, all dialects follow the same SSA rules and operation field format, which allows code analysis to be unified across different dialects.
MLIR maintains a list of core or contributed dialects.
The list is extensive, ranging from generic/low-level dialects (e.g., LLVM-IR/NVVM) to high-level dialects like linear algebra or tensor operations. Different dialects can be mixed in the MLIR.
To summarize, MLIR is a compiler eco-system, which encourages sharing and out-of-box transformation.

\subsection{MLIR in TensorFlow}
To date MLIR, 
been adopted only by TensorFlow, and Google is leading the development of both platforms.
As background for GEVO-ML, we next describe briefly how a DNN model is processed and represented in TensorFlow with respect to the MLIR.

TensorFlow's compiler is called XLA~\cite{xla}.  It compiles a DNN model using Google's IR, called High-level Operation (HLO). XLA performs target independent optimization on HLO before further translating into CPU code via LLVM-IR, Nvidia GPU code with NVVM, or via a private, unpublished IR for Google TPU. 

All the above representations, HLO, LLVM, NVVM, and other formats, can be expressed 
as a MLIR dialect. 
Figure~\ref{fig:2fc_mhlo} shows an example of how MLIR represents a TensorFlow model under the HLO dialect.

GEVO-ML is designed to interact with MLIR in the HLO dialect, mainly because this dialect is currently the most stable and has complete and precise documentation. 
However, GEVO-ML's interface can easily be expanded to other MLIR dialects as they become available.
\section{GEVO-ML Design}
\label{sec:Design}

In this paper, we propose GEVO-ML---an evolutionary computation tool for automatically searching for DNN model architectures and execution optimizations, focusing on MLIR.
GEVO-ML takes a MLIR program, a fitness function to optimize, and user-supplied datasets as inputs. The datasets serve as test cases for GEVO-ML. GEVO-ML seeks to maximize the fitness function by evolving and evaluating mutated program variants. 
As mentioned earlier, 
we demonstrate GEVO-ML for the TensorFlow HLO dialect in MLIR, as shown in Figure~\ref{fig:gevoml}. 

The initial population is formed by making copies and applying random mutations to the original MLIR program. By default, three mutations are applied to each individual in the initial generation. 
GEVO-ML uses the Non-dominated Sorting Genetic Algorithm (NSGA-II)~\cite{nsga-ii}, extending the DEAP implementation.
Each new generation of individuals is formed by ranking them according to the objectives, recombining individuals, applying mutation, 
comparing the new variants to a set of elites retained from the previous generation, and finally selecting the next generation.  The next few subsections describe how we implemented these search procedures for MLIR optimization. 

\begin{figure*}[t]
	\includegraphics[width=1\textwidth]{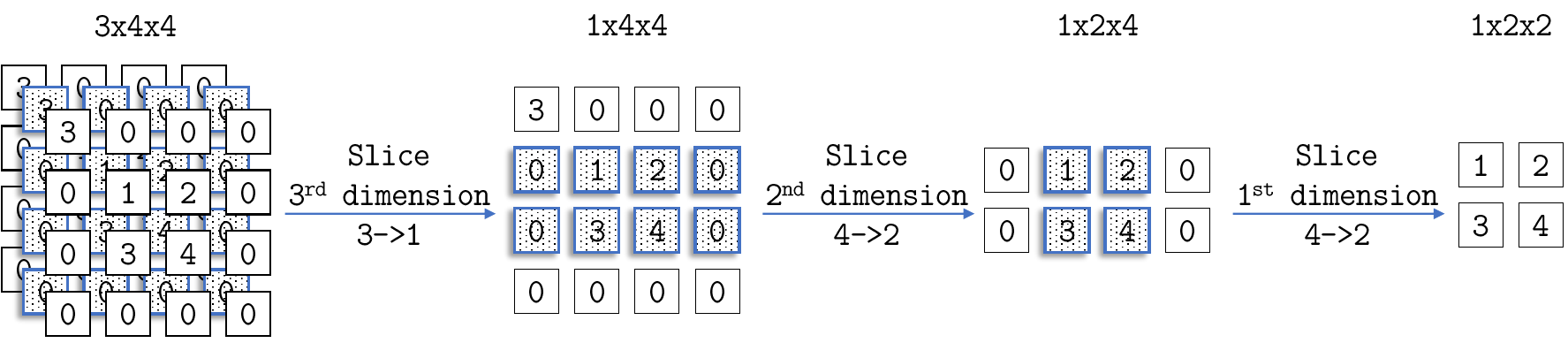}
	\caption{An example of GEVO-ML mutation: Shrink a tensor from size 3x4x4 to 1x2x2. The number of transitions shown in the figure corresponds to the number of MLIR operations required to reshape this tensor. }
	\label{fig:tensor_reshape} 
\end{figure*}

\subsection{Mutation}
\label{sec:design:mutation}

GEVO-ML uses two mutation operators, each of which modifies a MLIR instruction: Copy (copy a MLIR operation from one location in the program to another location); Delete (delete a MLIR operation). SSA enforces the requirement that the value of a variable can only be assigned once without being further modified. This makes the use-def chain explicit.
Mutations are highly likely to create invalid programs by violating this restriction and breaking variables in the use-def chain. GEVO-ML repairs the use-def chain by replacing invalid variable usage due to the mutation with other valid variables of the same types randomly.
In the HLO dialect, most variables are of type tensor, but operations tend to generate uniquely sized tensors, and tensors of different sizes are treated as different types. 
To repair the mutation, GEVO-ML shrinks or expands the selected tensor variable by dropping values from the tensor's edges or padding the tensor with value 1. Figure~\ref{fig:tensor_reshape} gives an example of a tensor mutation, which shrinks a tensor from size 3x4x4 to 2x2. 

The mutation operator selects one mutation type randomly (delete or copy) and applies the mutation to generate a new MLIR variant (with repairs as necessary).  After each mutation is applied, GEVO-ML immediately evaluates the edit against all test cases.  If it fails, the mutation operator selects another mutation until it finds a valid MLIR variant. 

\subsection{Crossover}
\label{sec:crossover}

For crossover, GEVO-ML uses a \textit{patch representation} in which 
an individual is represented as a list of edits to the original program.  We adopt this representation to maximize the chance of crossover producing a valid program. Here a program is a deep neural network. An alternative would be to recombine instructions from two different individuals, i.e., two neural networks. However, it is highly likely that recombining two random program slices will require many repairs to create a valid individual.  
GEVO-ML uses one-point messy crossover, which combines shuffle~\cite{shufflecrossover} and variable-length~\cite{Lee00variablelength} crossover operations.
GEVO-ML begins with two randomly selected individuals, concatenates the two lists of mutations (edits) in the patch representation; shuffles the sequence; and then randomly selects a location to cut the list back into two. GEVO-ML then reapplies each patch in sequence to the original GPU kernel and generates two new individuals. Although unusual, this strategy produces 
a wide diversity of recombinations from a minimal number of mutations. Mutations are relatively expensive in GEVO-ML due to the repair process.
Each new individual is then evaluated to test if the new combination of edits is valid, and we find that about 80\% of the time they are.
If not, GEVO-ML repeats the process until it finds a successful recombination. 

\subsection{Fitness Evaluation}
\label{sec:design:evaluation}
Individuals are evaluated according to the fitness objectives, e.g., runtime and model error.  Most earlier genetic improvement approaches require that an individual passes all of its input/output test cases exactly, or within a pre-specified error tolerance.  GEVO-ML requires only that
individuals execute successfully, and minimizes output error as one of the optimization objectives. 
This approach succeeds because ML applications can usually tolerate errors in the output, often in a vector composing the likelihood of each prediction category, as long as model accuracy can still be evaluated. 

The fitness objectives are to minimize MLIR program execution time and, at the same time, reducing the model error, namely, $argmin(time, error)$. 
Based on the specific ML task, fitness is evaluated either by
retraining the model on a given dataset and recording the training time and model error (\textit{training workloads}), or simply by passing dataset into the pre-trained model and recording the inference time and prediction error (\textit{prediction or inference workloads}). At the end of the search, the fittest individual is 
evaluated against a separate dataset unseen to GEVO-ML, to verify that the recorded time and error are consistent.

\subsection{Selection}

As in NSGA-II, GEVO-ML selects individuals according to multi-objective fitness criteria and reports the pareto frontier of individuals that best satisfy the two objectives.  
GEVO-ML retains the top 16 individuals at time $t$ and copies them unchanged to the population at time $t + 1$.
It then chooses the remainder of the population using tournament selection.

\section{Experimental Setup} 

GEVO-ML is developed in a modular fashion, with the main search framework implemented in Python using DEAP~\cite{deap}, which
interacts with a separate C++ program. The C++ program handles the MLIR parsing task and implements the MLIR mutation operations described in Section~\ref{sec:Design}. 

At the time of this work, the new, modular TensorFlow runtime system~\cite{tfrt}, that intends to take MLIR programs as an input, is under development. It does not support the end-to-end TensorFlow model execution completely. 
That is, the current TensorFlow compiler is not modular or flexible enough for third party programs to intercept the internal MLIR program, modify it, and then reinsert it for execution. However, the Google Intermediate Representation Execution Environment (IREE)~\cite{iree}, an experimental project, can execute TensorFlow models on edge/mobile devices and uses MLIR. THus, we build GEVO-ML for IREE as IREE has a functional MLIR execution runtime that can  
execute MLIR programs, independent of TensorFlow. 

We evaluate GEVO-ML on two neural network models, which are compatible with the IREE requirements. Despite its ability to execute external MLIR programs, IREE is under development and lacks support for many TensorFlow operations. 
Note, the MLIR IREE runtime is less performant than the native TensorFlow framework, which we expect to improve in the near future.
With these considerations in mind, we selected MobileNet~\cite{howard2017mobilenets} to evaluate GEVO-ML's ability to optimize model prediction task by minimizing the execution time of forward pass and maximizing model accuracy.  MobileNet is a highly efficient convolution neural network architecture. 
The weights of the model is retrieved from a pre-trained TensorFlow model. For model training, we chose a simple neural network with two fully-connected layers (denoted as 2fcNet in the rest of the paper). GEVO-ML is able to freely optimize both model forward pass and back-propagation pass. Stochastic gradient decent (SGD) is used as the training operation for this model. Currently, SGD on the fully-connected layer is the only functional model training workload supported by the IREE runtime. Training MobileNet requires SGD on the convolution layer, which is not yet supported. However, it is possible for GEVO to modifiy the SGD into other form during the search process.

\begin{table}
\centering
\caption{Model Parameters}
\label{tab:model_parameters}
    \begin{tabular}{| c | c | c |} 
    \hline
    & MobileNet & 2fcNet\\
    \hline
    \hline
    \makecell{Layer \\ composition} & \makecell{\textbf{17x} Depthwise- \\ Convolution \\ \textbf{35x} Standard- \\ Convolution \\ \textbf{52x} Batch Norm.\\ \textbf{1x} Average Pool \\ \textbf{2x} Fully- \\connected Layer} & \makecell{\textbf{2x} Fully- \\connected Layer} \\
    \hline
    \end{tabular}
\end{table}

We allocated a 48 hour wall-clock budget for GEVO-ML to optimize each model
on an Nvidia P100 GPU.  As mentioned earlier, the fitness function rewards both runtime and model error. 
CIFAR10~\cite{cifar10} is used for model prediction in MobileNet
where we use only the training data set (50,000 samples) to calculate model accuracy during the GEVO-ML run. The testing dataset (10,000 samples) is used post hoc to evaluate and verify model quality and execution time. Similarly, the MNIST~\cite{mnist} dataset (split into the 60,000 training and 10,000 testing samples) is used as the neural network training workload in 2fcNet due to the longer execution time and heavier computation requirement in model training.


\section{Experimental Results and Analysis}
\label{sec:result}

\begin{figure}
    \includegraphics[width=1\linewidth]{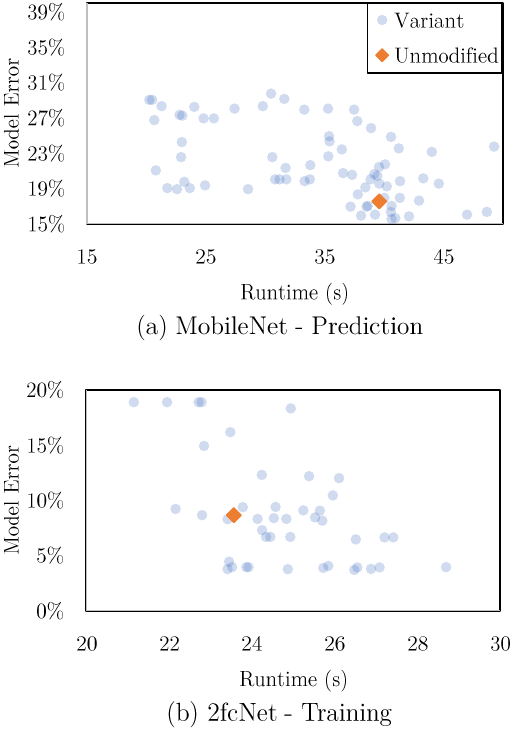}
	\caption{GEVO-ML result in Runtime/Model Error Pareto Front for (a) MobileNet - Prediction and (b) 2fcNet - Training. The orange diamond shows the original model whereas blue dots are the modification from original model generated by GEVO-ML.}
	\label{fig:paretofront} 
\end{figure}

Figure~\ref{fig:paretofront} shows the Pareto frontier results on the last generation of the GEVO-ML search for each model. Overall, GEVO-ML improves both the execution time performance and model quality---it improves the execution time by
by 90.43\% from 39.59 to 20.79 seconds for the prediction task (MobileNet with CIFAR10) and improves model accuracy by 4.88\% from 8.62\% to 3.74\% for the training task (2fcNet with MNIST). 
For both models, we evaluate model accuracy using the training dataset and examined the testing dataset for the model variants GEVO-ML discovered. In MobileNet (Prediction), no testing accuracy improvement was observed, and if we can tolerate a 2\% reduction in testing accuracy, then we achieve 90\% execution time performance improvement. In 2fcNet (Training), we obtain 5\% training accuracy, which is preserved when we evaluate on the testing data.
In the following subsections, we examine how these performance/accuracy improvements were achieved. 

\subsection{Mutation analysis: Model Prediction in MobileNet}
\label{sec:result-mobilenet}
In the MobileNet experiments, we found three key mutations that contributed to improved  performance/accuracy trade-offs: 
\begin{itemize}
  \item Replacing the $\gamma$ value in one Batch Normalization (BN) layer with the $\gamma$ value in  its prior BN layer
  \item Removing the bias term from the last fully connected layer
  \item Removing the last convolution layer
\end{itemize}

It is challenging to interpret exactly how these mutations reduce the model's accuracy, but the significant 90\% performance improvement can be explained to some extent. The three key mutations are epistatic and work together synergistically to reduce runtime. When considered individually, none of the mutations has a significant impact on performance. For example, one mutation removes the last convolution layer which one might expect would have large impact. However, there are 52 convolution operations in total, and the last layer certainly does not contribute to a half of the runtime overhead.
Taken alone, this mutation does not have large performance effect, but when applied in conjunction with the other two mutations, the 90\% execution time improvement is obtained.  Although we were unable to explain exactly how these mutations work, we speculate that 
the mutation combinations influence the multiple transformation passes and introduce low-level optimizations in IREE.  
This example shows how GEVO-ML can discover implicit, non-obvious optimization opportunities in these highly complex software systems and provide the code variant as an option for programmers to investigate. 
\begin{figure}
    \includegraphics[width=1\linewidth]{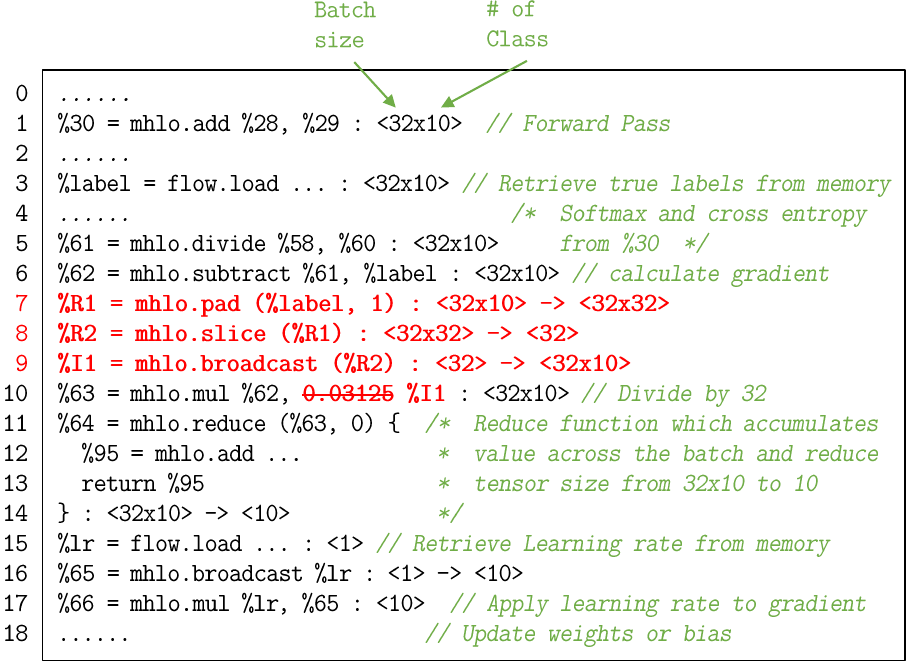}
	\caption{GEVO-ML optimization of the training workload. The code snippet shows the original MLIR
	code, which calculates a gradient based on the batch size. The highlighted code (Line 7-9 and part of Line 10) shows the mutation that led to 4.88\% training accuracy increase.}
	\label{fig:mod_analysis_train} 
\end{figure}

\subsection{Mutation analysis: Model training in 2fcNet}
\label{sec:result-2fc}
In contrast with the previous example, a 4.88\% accuracy improvement was achieved with a single mutation in the model training workload. At a high level, primary impact of the mutation increases the degree of gradient, leading the model to update weights more aggressively. 

Figure~\ref{fig:mod_analysis_train} shows the GEVO-ML mutation (highlighted in red), inside the process that updates model weights and bias. In SGD with mini-batch, the model weights are updated using gradient from each example in one 'mini-batch,' which is multiplied with the learning rate. The gradient of a mini-batch is first retrieved as the difference between the true label and the model prediction value from the forward pass (Lines 1-6). To average out the gradient in the mini-batch, the accumulation (the reduce function in Lines 11-14) and the divide operation (the multiplication with 1/32 in Line 10 where 32 is the batch size) are integrated to the calculation. The mean of the gradient is then multiplied with the leaning rate and used to change the weight and bias of the model (Lines 15-18) . 

The single GEVO-ML mutation is shown in Line 9. GEVO-ML copies a broadcast operation from another location in the program, connects the \textit{\%label} as input and inserts the output of the newly copied operation into the next operation (Line 10), replacing the value 0.0325 (seen at the top of the figure in Line 7). However, since the copied broadcast operation has the input type signature (tensor<32>) which is mismatched with the intended input variable \%label (tensor<32x10>), GEVO-ML performs the repair process (Section~\ref{sec:design:mutation}), with two additional operations that modify the tensor into a compatible shape (pad and slice operations in Lines 7-8). After the tensor containing labels is reshaped, it is filled mostly by value '1', and only one label vector remains in the center of the tensor. The average value within the reshaped tensor is certainly larger than  
the value 0.03125 used in the unmodified model, resulting in a larger gradient value in general. Thus, we infer that the accuracy improvement is achieved with more aggressive training updates through larger gradients.
Although most end users cannot directly control or scale the gradient, as GEVO-ML did, increasing the learning rate, which is applied in Line 15, can also enlarge the gradient and achieve a similar effect. We verified this assumption by increasing the learning rate from 0.01 to 0.3 and achieve comparable accuracy improvement.


\section{Discussion, Challenge, and Future Opportunities}

In GEVO-ML, mutations either copy or delete existing HLO operations and then connect variables. However, many HLO operations provide an additional mode in their attribute field, which is statically assigned, e.g., the kernel size of a convolution operation. This is a tempting target as an additional mutation operator. We did not implement this particular operator in GEVO-ML because it would expand the search space intractably.   

Prior works in genetic improvement of software have sometimes supported mutations of constant numbers indirectly~\cite{GP4AutoSWrepair, schulte2015repairing}. 
A typical approach is to rely on existing numbers in the program, e.g., stored as variables, and allow mutation to manipulate them through arithmetic instructions. This approach is currently not feasible in MLIR because the attribute field can accept only predefined constant values, and no dynamic assignment through in-program variables is allowed. Thus, we leave it as future work for GEVO-ML.

GEVO-ML contributes new mutation operators, which change the size of the tensor variables, which are the primary variable type in the HLO dialect.  Our motivation for this mutation operator was to enhance GEVO-ML's ability to mutate HLO programs successfully. 
We acknowledge that changing the tensor type by dropping or padding values could change program semantics dramatically. It is an interesting future avenue to understand the effect of tensor size mutation on the behavior of ML models more extensively. 
We also expect that, when the HLO dialect is lowered onto stable dialects such as linear algebra or affine
dialects, this issue will be mitigated
because tensor operations will then be decomposed into loops and fundamental arithmetic operations on integer or floating point numbers. 

Another limitation on GEVO-ML's implementation comes from the immature MLIR eco-system. Only TensorFlow has actively adopted  MLIR, although Google has some other internal implementations. Moreover, the TensorFlow modular runtime system is still under development.
We expect this situation to change quickly now that MLIR is upstream in the LLVM family, which means that LLVM-IR can be treated as a subset of MLIR. We expect GEVO-ML to reveal more unseen optimization opportunities, as MLIR continues to expand its reach into many domains, including more fundamental and general usage and additional back-end device support, with integration beyond deep neural networks. 
Finally, given the drastically-rising cost of machine learning~\cite{henderson2020towards, strubell2019energy, bender2021dangers, amodei2018ai}, we believe automatic optimization tools, such as GEVO-ML, are instrumental to accelerate the performance and efficiency optimization for deep learning.

\section{Conclusion}

GEVO-ML is an automatic EC tool for cross-layer optimization of ML software. We demonstrate GEVO-ML running on a production level ML framework using the MLIR as its program representation. Although the demonstration uses a single MLIR dialect, GEVO-ML identifies optimizations for individual parameters (changing the learning rate), model architecture management (removing a convolution layer), and we speculate that it finds indirect low-level implementation optimizations by combining operation manipulations in subtle ways.
As ML models continue to grow in size and complexity and ML infrastructures continue to mature, we hope that evolutionary computation as illustrated with GEVO-ML will play a key role in continuing to improve the design and execution characteristics of ML.

\bibliographystyle{ACM-Reference-Format}
\bibliography{ref}

\end{document}